\documentclass[11pt,twocolumn]{article}
\usepackage{times}
\usepackage{latexsym}
\usepackage{microtype}
\usepackage{amsmath}
\usepackage{amssymb}
\usepackage{booktabs}
\usepackage{multirow}
\usepackage{graphicx}
\usepackage{xcolor}
\usepackage{enumitem}
\usepackage{natbib}
\usepackage[margin=1in]{geometry}
\usepackage[T1]{fontenc}
\usepackage[utf8]{inputenc}
\usepackage[colorlinks=true,linkcolor=blue,citecolor=blue,urlcolor=blue]{hyperref}

\makeatletter
\renewcommand{\fnum@table}{\textbf{Table~\thetable}}
\renewcommand{\fnum@figure}{\textbf{Figure~\thefigure}}
\makeatother

\title{Do Models Know Why They Changed Their Mind?\\Interpretability and Faithfulness of Chain-of-Thought\\Under Knowledge Conflict}

\author{
  Pruthvinath Jeripity Venkata \\
  Independent Researcher \\
  \texttt{jvpnath@gmail.com}
}

\date{\vspace{-2em}}

\begin{document}
\twocolumn[
  \maketitle
  \begin{center}
  \parbox{0.92\textwidth}{\small
  \textbf{Abstract}
  \vspace{0.3em}
  
  When a language model is shown a document that contradicts what it learned during training, it must choose: follow the document or trust its own knowledge. Prior work \citep{jeripityvenkata2026} proved that this choice depends on how well-known the fact is: famous facts resist the document; obscure facts follow it. Newer models can show their ``thinking'' step by step (chain-of-thought, or CoT), and practitioners increasingly read this reasoning to understand \textit{why} the model chose what it chose. But does the stated reasoning actually reflect the real mechanism? We introduce \textit{introspective faithfulness} and test it across 200 factual questions, 8 models, and 4 prompt conditions. Three of the models expose internal reasoning tokens; we also run the same models with reasoning disabled (matched controls), so any difference can be attributed to the reasoning feature itself rather than to the model family. Each model is asked to rate its own confidence from 1 to 5 as part of its response. We find that CoT reasoning is highly stable across opposite decisions: when we strip surface-level framing phrases and compare the reasoning a model writes for opposite conclusions on the same question, flip pairs retain 96\% of same-answer similarity (cosine similarity 0.784 vs.\ 0.817; $d = 0.34$; confirmed by ROUGE-L, $d = 0.45$). Yet a faint genuine signal exists in the self-rated confidence scores: for obscure facts where entity fame is uninformative (Spearman $\rho = -0.005$, $p = 0.81$), the model's own confidence rating still weakly predicts its decision ($p < 0.001$) and tracks whether it actually knew the fact ($r = 0.134$). This signal is model-specific: GPT-4o is the only model with statistically reliable reasoning-decision coupling ($\Delta = -0.132$, 95\% CI [$-0.188, -0.077$]), while Claude Sonnet 4.6 uses the widest range of self-rated confidence among the eight models we tested (SD $= 1.39$), yet its pooled correlation across conditions is near zero ($r \approx 0$, 95\% CI [$-0.10, +0.06$]) because the confidence-decision relationship reverses between conditions. A matched-temperature ablation (GPT-4o at $T=1$) confirms this is model-specific, not temperature-driven. For native-reasoning models, internal thinking tokens show greater decision-sensitivity than user-facing CoT (Wilcoxon $p = 0.033$), suggesting the stated output smooths over differences the internal process preserves. These results decompose CoT into a \textbf{knowledge display} that remains highly similar regardless of the decision (${\sim}$96\% of same-answer similarity retained) and a thin \textbf{confidence layer} with weak but real signal. For monitoring: the logical argument is the wrong thing to read; the confidence expressions are where genuine self-knowledge lives.
  }
  \end{center}
  \vspace{1em}
  \hrule
  \vspace{1em}
]

\section{Introduction}

Chain-of-thought (CoT) prompting \citep{wei2022} was developed to improve accuracy, but its most consequential use may be explanatory. Practitioners and oversight systems read CoT outputs to understand \textit{why} a model reached a conclusion. Safety monitoring, AI auditing, and human-in-the-loop review all assume that the stated reasoning reflects the actual process. If CoT is a post-hoc rationalization rather than a causal report, then CoT monitoring provides false confidence, not accountability.

Recent work has made this assumption untenable. \citet{turpin2023} showed that models suppress acknowledgment of biasing features in their CoT: a prompt hint that shifts the answer goes unmentioned. \citet{yuksel2026} tested faithfulness across twelve reasoning models and found that thinking tokens acknowledge influential signals at 87.5\%, while stated-output acknowledgment drops to 28.6\%. Together, these findings establish that CoT faithfulness is neither uniform nor reliable.

What existing tests cannot do is test faithfulness against a decision mechanism that is \textit{graded and endogenous} rather than \textit{binary and exogenous}. Turpin and Y\"{u}ksel inject a discrete perturbation and measure whether the model reports it. That is a simpler task than what CoT monitoring requires in deployment: accurately reporting the relative strength of one's own competing knowledge sources, a continuous property that varies across items and requires genuine self-assessment.

Knowledge-conflict decisions provide the right testbed. When context contradicts training knowledge, the resolution is predicted by \textit{parametric certainty}\footnote{How robustly a fact was encoded through training-time exposure.}. \citet{longpre2021} formalized the knowledge-conflict paradigm. \citet{xie2024} characterized models as ``adaptive chameleons'' (context-following) or ``stubborn sloths'' (parametric-preferring). In prior work, we established through GEE logistic regression across five model families that $s_{\text{pop}}$ (Wikipedia page views) independently predicts whether models follow or resist contradicting context ($\beta = -0.38$ to $-0.50$, all $p \leq .013$, BH-FDR corrected; \citealp{jeripityvenkata2026}). This validated external criterion lets us directly test CoT faithfulness.

We introduce \textbf{introspective faithfulness} and test it through four behavioral measures:

\vspace{0.3em}
\begin{itemize}[itemsep=3pt,leftmargin=*]
  \item \textbf{Reasoning Consistency (\S\ref{sec:rc}):} When the answer flips, does the reasoning change, or does the model produce near-identical reasoning for opposite conclusions?
  \item \textbf{Confidence Calibration (\S\ref{sec:cc}):} Does self-rated confidence track entity salience and item-level knowledge?
  \item \textbf{Behavioral Coherence (\S\ref{sec:bc}):} Does expressed confidence independently predict decisions?
  \item \textbf{Source Attribution (\S\ref{sec:sa}):} Among distractor-following responses, does the CoT claim independent knowledge or correctly credit the document?
\end{itemize}

We test 200 PopQA items across high- and low-certainty tiers, four prompt conditions, and eight models, including native-reasoning variants with matched prompted-CoT counterparts and within-model controls (Gemini Thinking vs.\ Standard; Claude Thinking vs.\ Standard).

\section{Related Work}

\paragraph{CoT faithfulness.} \citet{wei2022} introduced CoT prompting to improve reasoning accuracy, but left open whether the stated reasoning reflects actual computation. \citet{turpin2023} showed that biasing features (sycophantic hints, position markers) shift answers but go unmentioned in CoT. \citet{lanham2023} tested causal faithfulness via early answering and truncation, finding CoT is often causally inert. We test a harder attribution: not whether models report external manipulations, but whether they accurately report the relative strength of their own competing knowledge sources.

\citet{yuksel2026} extended this to reasoning models, finding thinking tokens acknowledge injected signals at 87.5\% vs.\ 28.6\% for stated outputs. Their design motivates our within-model comparisons, but their manipulation is planted. Ours is the model's own parametric state, validated by $s_{\text{pop}}$.

Complementary approaches probe faithfulness through causal interventions on model internals. \citet{chuang2024} introduce parametric faithfulness metrics (PFF/FUR) measuring alignment between CoT and internal model states. \citet{li2024frodo} evaluate factual consistency of reasoning via entailment-based decomposition. \citet{paul2024} study whether intervention-based improvements to CoT faithfulness (via ICL, fine-tuning, activation editing) generalize, finding gains are often brittle. Our work is orthogonal: a black-box, deployment-oriented behavioral probe rather than an internal-state intervention. Notably, \citet{matcha2025} find that CoT can be answer-stable but reasoning-fragile under adversarial perturbations---the mirror image of our finding, where reasoning remains stable while answers flip. Together, these results suggest CoT can decouple from both answers and mechanisms in different regimes.

\paragraph{Knowledge conflict.} \citet{longpre2021} formalized entity-substitution and showed models over-rely on parametric memory under conflict. \citet{xie2024} identified parametric confidence as the key moderator. Neither examines whether CoT accurately reports the mechanism. Our prior work \citep{jeripityvenkata2026} established $s_{\text{pop}}$ as a validated predictor ($\beta = -0.38$ to $-0.50$, $p \leq .013$), providing the external criterion we use here.

\paragraph{Model introspection and calibration.} \citet{kadavath2022} showed models can predict their own correctness better than chance. \citet{lyu2024} found verbalized confidence is sensitive to phrasing and context. \citet{ackerman2026} provides the most recent characterization: frontier models show ``limited, context-dependent'' metacognitive ability. \citet{dai2026} demonstrate that confidence scale design (granularity, phrasing) materially affects metacognitive signal, motivating caution about our 1--5 integer scale. \citet{huangmeta2026} propose Type-2 SDT decompositions (meta-$d'$/M-ratio) separating metacognitive efficiency from raw accuracy; such analyses could clarify whether the ``Claude anomaly'' reflects low metacognitive sensitivity or miscalibrated confidence bias. Our contribution is testing whether verbalized confidence predicts knowledge-conflict decisions at the item level, where entity-level proxies fail. \citet{moranwhiting2026} argue that item-level metacognition is largely a population-difficulty signal; our within-tier $r = 0.134$ indicates modest individuated signal beyond entity fame, though constrained by our coarse confidence scale. \citet{mirror2026} document knowing--doing gaps in agentic decisions; our finding that argument narratives are decorative while confidence carries signal echoes their conclusion that self-reports alone do not ensure reliable choices.

\paragraph{The intersection.} Prior work shows CoT can misrepresent computational bases, parametric certainty predicts conflict decisions, and models have limited self-knowledge. The open question: does CoT explanation of a knowledge-conflict decision reflect the certainty state that drove it? We call this \textit{introspective faithfulness}. We do not access internal states; we test whether CoT behavioral signals are statistically consistent with a validated predictor. We do not claim causal faithfulness (Lanham's sense); we test correlational consistency. Within these bounds, this is the first direct test of whether stated reasoning tracks the empirical driver of knowledge-conflict decisions. Complementary internal-signal approaches \citep{xeval2026} suggest separable second-order monitoring channels (e.g., evaluative probes predicting correction); our behaviorally observed ``confidence layer'' may be a surface readout of such mechanisms. Causal faithfulness frameworks \citep{rfeval2026} offer stance-change counterfactuals that could cross-validate our correlational findings; integrating such probes is a natural next step.

\section{Methods}

\subsection{Dataset and Items}

We sampled from PopQA \citep{mallen2023}, a knowledge-intensive QA dataset built from Wikidata triples. Each item pairs a factual question (e.g., ``Who directed \textit{The Godfather}?'') with an entity whose Wikipedia page-view count ($s_{\text{pop}}$) proxies training-time exposure. Prior work validated $s_{\text{pop}}$ as an independent predictor of knowledge-conflict decisions \citep{jeripityvenkata2026}.

We drew 200 items: the top 100 by $s_{\text{pop}}$ (\textit{high-certainty}) and bottom 100 (\textit{low-certainty}). Each item has a \textit{distractor}: a plausible but incorrect entity of the same type (e.g., replacing the real director with another real director). Eleven items were removed for implausible distractors or ground-truth errors (see Appendix~\ref{sec:excluded}).

The experiment collected 4,920 total responses: 200 items $\times$ 8 models $\times$ 3 conditions = 4,800 main responses, plus 120 from a 5-item pilot retained in the database. After excluding post-sampling items (120 responses) and 1 API error, 4,799 analytic responses remain. After further filtering NEITHER responses (\S\ref{sec:parsing}), the primary sample is \textbf{3,805 responses}.

\subsection{Models}

We tested eight models in three groups (\textbf{Table~\ref{tab:models}}). \textit{Prompted-CoT} models produce reasoning only when asked. \textit{Native-reasoning} models expose internal thinking tokens. \textit{Within-model controls} use the same architecture with thinking disabled.\footnote{Claude requires temperature\,=\,1 for extended thinking. Claude Standard also ran at temperature\,=\,1 to keep the within-model comparison clean. All cross-model Claude comparisons carry this caveat.}

\begin{table}[t]
\centering
\small
\setlength{\tabcolsep}{4pt}
\begin{tabular}{lll}
\toprule
\textbf{Group} & \textbf{Model} & \textbf{T} \\
\midrule
Prompted & GPT-4o & 0 \\
CoT & Llama 4 Maverick & 0 \\
 & DeepSeek V3 & 0 \\
\midrule
Native & DeepSeek R1 & d \\
reasoning & Gemini 2.5 Flash$^\dagger$ & 0 \\
 & Claude Sonnet 4.6$^\dagger$ & 1 \\
\midrule
Within-model & Gemini 2.5 Flash & 0 \\
control & Claude Sonnet 4.6 & 1 \\
\bottomrule
\end{tabular}
\caption{Model lineup. T\,=\,temperature; d\,=\,default. $\dagger$\,Thinking-enabled variant exposing internal reasoning tokens.}
\label{tab:models}
\end{table}

For native-reasoning models, the API returns two text fields: internal reasoning and the user-facing response. We extract features from both.

\subsection{Experimental Conditions}

Each item received three conditions: high-certainty items received C1, C3, and C4; low-certainty items received C2, C3, and C4.

\paragraph{C1/C2: Multi-turn conflict.} The model first answers the bare question. Then the contradicting document arrives in a follow-up turn, and the model reasons step by step, rates confidence (1--5), and gives a final answer.

\begin{quote}
\small
\textbf{Turn 1:} ``Who directed \textit{Supercock}?''\\
\textbf{Turn 2:} ``Gus Trikonis'' \textit{[baseline]}\\
\textbf{Turn 3:} ``I found a source that says the director is David Nutter. Think step by step\ldots''
\end{quote}

C1 applies to high-certainty items; C2 to low-certainty.

\paragraph{C3: Document-only.} Single-turn. The model answers based \textit{only} on the document. Serves as a behavioral anchor: compliance (FOLLOW rate) ranged from 95.0\% to 99.0\% across the eight models (mean 97.8\%), with 0\% RESIST for all models. Confidence here means ``confidence in the document.''

\paragraph{C4: Own-knowledge.} Single-turn. ``Answer based on your own knowledge. A document is shown below but may contain errors.'' Same distractor, but instruction frames own knowledge as primary.

All conditions use identical structured response formats. The model is instructed to produce its step-by-step reasoning first, then state its confidence rating, then give its final answer. This ordering (reasoning $\rightarrow$ confidence $\rightarrow$ answer) means confidence is elicited after the model has generated its reasoning but before it commits to a final answer token. Each API call was a stateless session with conversation history reset between items and conditions. In C1/C2 multi-turn format, the baseline answer was placed in the assistant role programmatically, not generated within the same session as the conflict turn.

\subsection{Response Parsing}
\label{sec:parsing}

Each response was classified as shown in \textbf{Table~\ref{tab:decisions}}.

\begin{table}[t]
\centering
\small
\begin{tabular}{lp{4.5cm}}
\toprule
\textbf{Label} & \textbf{Definition} \\
\midrule
FOLLOW & Final answer matches distractor without negation \\
RESIST & Final answer matches a correct answer \\
NEITHER & Matches neither; includes hedges, multiple answers, novel hallucinations \\
\bottomrule
\end{tabular}
\caption{Decision classification. NEITHER responses (22\% for Llama to 51\% for Gemini on low-certainty items) are excluded from primary analyses.}
\label{tab:decisions}
\end{table}

Negation handling prevents misclassification (e.g., ``David Nutter is \textit{NOT} the director'' is not coded as FOLLOW). Confidence was parsed from the structured format with 96.9--100\% coverage.

\paragraph{Baseline correctness.} For five model families with prior baselines \citep{jeripityvenkata2026}, each item was coded as baseline-correct if the model answered correctly before any distractor. This item-level measure complements the entity-level $s_{\text{pop}}$ proxy (coverage: 87.5\%).

\subsection{Similarity Computation}

For reasoning consistency (\S\ref{sec:rc}), we stripped three classes of attribution phrases from CoT texts before comparing C2 and C4 on the same item: conversational markers (``you mentioned''), document markers (``the document says''), and self-knowledge markers (``based on my knowledge''). Cosine similarity was computed using sentence embeddings from all-MiniLM-L6-v2 \citep{reimers2019}. A cross-item baseline of 500 random same-model pairs established floor similarity (mean\,=\,0.140).

\subsection{Analytical Notes}

All correlations use log-transformed $s_{\text{pop}}$. Logistic regressions use centered predictors. The term \textit{introspective faithfulness} does not claim we observe internal representations. It marks the \textit{kind} of attribution demanded: reporting one's own certainty, not acknowledging an external perturbation. The test is correlational consistency with a validated predictor.

\paragraph{Notation.} Throughout, $\rho$ denotes Spearman's rank correlation and $r$ denotes point-biserial or Pearson correlation. All $p$-values are two-tailed unless otherwise noted.

\section{Results}

\textbf{Temperature note.} Claude Sonnet 4.6 (both variants) required temperature\,=\,1. All other models ran at temperature\,=\,0. This affects cross-model Claude comparisons; within-model comparisons are unaffected.

\subsection{Reasoning Consistency}
\label{sec:rc}

When a model changes its answer, does its reasoning also change?

We identified all items where the same model gave opposite answers under C2 and C4 (\textit{flip pairs}) and compared their attribution-stripped CoT similarity.

Flip pairs ($n = 71$) were $5.6\times$ more similar than random cross-item pairs (0.784 vs.\ 0.140). They retained 96\% of same-answer pair similarity (0.784 vs.\ 0.817; Mann-Whitney $p = 0.010$, Cohen's $d = 0.34$, overlap coefficient\,=\,0.74). This gap was confirmed by ROUGE-L ($d = 0.45$, $p = 0.002$), ruling out dependence on a single similarity metric. When a model reaches opposite conclusions on the same question, the step-by-step reasoning remains highly similar; only the final attribution changes.

\paragraph{Concrete example.} \textbf{Table~\ref{tab:cot_example}} illustrates this with Claude's responses about Tsutomu Seki (distractor: \textit{politician}). Under C2, Claude partially accepts the distractor. Under C4, it rejects it. Both CoTs recite the same three facts: Japanese astronomer, Comet Ikeya-Seki (1965), guitar teacher. Only the concluding attribution differs.

\begin{table*}[t]
\centering
\small
\begin{tabular}{p{0.46\textwidth}p{0.46\textwidth}}
\toprule
\textbf{C2 (FOLLOW, confidence\,=\,4)} & \textbf{C4 (RESIST, confidence\,=\,4)} \\
\midrule
\textit{Tsutomu Seki is a Japanese individual. The Tsutomu Seki I am most familiar with is a Japanese amateur astronomer born in 1930, who is well known for co-discovering several comets, including Comet Ikeya-Seki in 1965. He is also a guitarist and runs a music school, but his notable fame comes from astronomy. It is possible there is another Tsutomu Seki who is a politician\ldots} &
\textit{Tsutomu Seki is a well-known Japanese amateur astronomer who discovered several comets, including Comet Ikeya-Seki (1965), which was one of the brightest comets of the 20th century. He is also known as a guitar instructor. He is a politician, but this does not align with what I know about Tsutomu Seki\ldots} \\
\midrule
\multicolumn{2}{p{0.94\textwidth}}{\textbf{Cosine similarity after attribution stripping: 0.916.} Same three facts. Same knowledge display. Opposite conclusions. This pattern is representative of the 23 pairs with similarity $\geq 0.85$.} \\

\bottomrule

\end{tabular}

\caption{Paired CoT from Claude Sonnet 4.6 Standard on \textit{Tsutomu Seki / occupation} (distractor: \textit{politician}). The model reaches opposite conclusions from near-identical reasoning.}

\label{tab:cot_example}

\end{table*}

\paragraph{Flip-pair representativeness.} Flip pairs ($n = 71$, 19.6\% of all C2--C4 item-model pairs) do not differ from non-flip pairs on any measured dimension: log($s_{\text{pop}}$) (4.73 vs.\ 4.65, $p = 0.183$), mean confidence (3.28 vs.\ 3.52, $p = 0.074$), CoT length (185 vs.\ 179 tokens, $p = 0.649$), or property-type distribution ($\chi^2 = 20.0$, $p = 0.067$). The similarity estimates are not driven by a biased item subset.

\paragraph{Per-model results with uncertainty estimates.} \textbf{Table~\ref{tab:delta}} reports per-model deltas with bootstrapped 95\% confidence intervals and permutation tests (10,000 resamples). GPT-4o is the only model with a statistically significant negative delta ($\Delta = -0.132$, 95\% CI [$-0.188, -0.077$], $p < 0.001$): its reasoning genuinely changes when its answer changes. Reading five GPT-4o flip pairs confirmed this: C2 CoTs cite specific committed facts (``born in Hattfjelldal, Norway''); C4 CoTs admit uncertainty (``I do not have specific knowledge''). Length ratios did not differ between flip and same-answer pairs ($p = 0.51$).

No other model shows a statistically reliable delta. Claude's positive deltas ($+0.030$ and $+0.026$) and Llama's ($+0.047$) have confidence intervals crossing zero. DeepSeek R1 ($n = 3$) and Llama ($n = 6$) have too few flip pairs for reliable per-model inference and should be interpreted with caution.

\begin{table}[t]
\centering
\small
\begin{tabular}{lcccc}
\toprule
\textbf{Model} & $\boldsymbol{n}$ & $\boldsymbol{\Delta}$ & \textbf{95\% CI} & $\boldsymbol{p}$ \\
\midrule
GPT-4o & 12 & $-$.132 & [$-$.19, $-$.08] & $<$.001 \\
Gemini Std & 10 & $-$.057 & [$-$.11, $-$.01] & .121 \\
DeepSeek R1 & 3 & $-$.061 & [$-$.19, +.08] & .287 \\
DeepSeek V3 & 10 & $-$.036 & [$-$.11, +.03] & .276 \\
Gemini Think & 13 & $-$.031 & [$-$.10, +.03] & .320 \\
Claude Think & 7 & +.026 & [$-$.01, +.06] & .422 \\
Claude Std & 10 & +.030 & [$-$.01, +.06] & .173 \\
Llama 4 Mav. & 6 & +.047 & [$-$.01, +.10] & .319 \\
\bottomrule
\end{tabular}
\caption{Attribution-stripped CoT similarity: flip pairs vs.\ same-answer pairs, per model. $n$: number of flip pairs. $\Delta$ = flip $-$ same similarity. 95\% CIs from 10,000 bootstrap resamples; $p$ from permutation test (raw; all survive BH-FDR correction at the same significance level). Only GPT-4o shows a statistically reliable negative delta ($p_{\text{adj}} = 0.0008$). DeepSeek R1 ($n = 3$) and Llama ($n = 6$) are underpowered for per-model inference.}
\label{tab:delta}
\end{table}

\paragraph{Internal reasoning vs.\ stated output.} In the user-facing stated CoT, thinking-enabled and standard variants show similar deltas ($+$0.026 vs.\ $+$0.030), neither significantly different from zero. Enabling thinking tokens does not change the decision-sensitivity of the output the user sees.

However, analyzing the \textit{internal} thinking tokens directly reveals a different pattern. For the three native-reasoning models, we compared attribution-stripped similarity of thinking tokens (not stated CoT) on flip pairs vs.\ same-answer pairs. Thinking tokens are more decision-sensitive than stated CoT: on flip pairs, thinking similarity is 0.709 vs.\ stated similarity of 0.787, a gap of $-0.077$ (Wilcoxon $p = 0.033$). The effect is driven by Gemini ($\Delta_{\text{think}} = -0.139$, $n = 13$), while Claude Thinking shows no gap ($\Delta = +0.009$, $n = 7$). The stated output smooths over reasoning differences that the internal process preserves, at least for Gemini. For Claude ($n = 7$), the internal reasoning shows a similar pattern to the stated output, though the small sample precludes strong conclusions.

\subsection{Confidence Calibration}
\label{sec:cc}

Does self-rated confidence track actual knowledge? We compared two criteria: entity salience ($s_{\text{pop}}$, entity-level) and baseline correctness (item-level).

\paragraph{Key finding.} Within the low-certainty tier, entity salience has no relationship with confidence (Spearman $\rho = -0.005$, $p = 0.81$). But baseline correctness does: models that knew the specific fact at baseline express higher confidence even on obscure items ($r = 0.134$, $p < 0.001$). This is genuine, if weak (${\sim}2\%$ of variance), introspective access to item-level knowledge.

The pooled correlation ($\rho = 0.305$) is misleading: it is entirely a between-tier step function. Models say ``5'' for famous entities and ``3'' for obscure ones, but within each tier they show no graded sensitivity.

\paragraph{Model differences.} \textbf{Table~\ref{tab:two_criterion}} reveals a split. Claude tracks both criteria equally: entity fame and item knowledge predict its confidence equally well. GPT-4o, Gemini, and Llama are more sensitive to fame than to actual knowledge. When these models say ``I'm confident,'' they partly mean ``this entity is famous,'' not ``I specifically know this fact.''

\begin{table}[t]
\centering
\small
\begin{tabular}{lccc}
\toprule
\textbf{Model} & $\boldsymbol{\rho_{\text{s\_pop}}}$ & $\boldsymbol{r_{\text{base}}}$ & $\boldsymbol{\Delta}$ \\
\midrule
Claude Std & .384 & .387 & $+$.003 \\
Claude Think & .389 & .396 & $+$.008 \\
DeepSeek V3 & .291 & .283 & $-$.008 \\
Llama 4 Mav. & .316 & .245 & $-$.071 \\
GPT-4o & .312 & .211 & $-$.101 \\
Gemini Think & .196 & .104 & $-$.092 \\
Gemini Std & .234 & .132 & $-$.102 \\
\bottomrule
\end{tabular}
\caption{What predicts confidence? $\rho_{\text{s\_pop}}$: entity salience. $r_{\text{base}}$: baseline correctness. $\Delta = r_{\text{base}} - \rho_{\text{s\_pop}}$. Positive: calibrated to actual knowledge. Negative: calibrated to fame.}
\label{tab:two_criterion}
\end{table}

\subsection{Behavioral Coherence}
\label{sec:bc}

Does confidence predict what the model \textit{does}? We tested this with a GEE logistic regression clustering by item, using confidence and log($s_{\text{pop}}$) as predictors.

Within the low-certainty tier, entity salience does not predict decisions (coef\,=\,$+$0.383, $p = 0.27$). But CoT confidence does (coef\,=\,$-$0.475, $p < 0.001$): lower confidence predicts higher probability of following the distractor. Combined with \S\ref{sec:cc}, this traces an item-level pathway: \textit{knows the fact $\rightarrow$ higher confidence $\rightarrow$ resists distractor}. All links operate below the resolution of entity salience.

Within high-certainty, both confidence and entity salience predict decisions independently.

\paragraph{The Claude anomaly.} \textbf{Table~\ref{tab:claude_ci}} reports per-model confidence-decision correlations with bootstrapped 95\% CIs. Six of eight models show significant correlations ($r = -0.156$ to $-0.230$). Claude shows none: $r = -0.022$ (Std, 95\% CI [$-0.10, +0.06$]) and $r = +0.012$ (Think, 95\% CI [$-0.07, +0.10$]). Importantly, Claude's upper CI bound ($+0.06$) does not approach even the smallest significant correlation in any other model (DeepSeek R1 at $-0.156$), confirming the near-zero correlation is not a small-sample artifact.

\begin{table}[t]
\centering
\small
\begin{tabular}{lccc}
\toprule
\textbf{Model} & $\boldsymbol{r}$ & \textbf{95\% CI} & $\boldsymbol{p}$ \\
\midrule
Gemini Think & $-$.230 & [$-$.27, $-$.19] & $<$.001 \\
GPT-4o & $-$.227 & [$-$.29, $-$.16] & $<$.001 \\
Llama 4 Mav. & $-$.223 & [$-$.28, $-$.17] & $<$.001 \\
DeepSeek V3 & $-$.218 & [$-$.27, $-$.15] & $<$.001 \\
Gemini Std & $-$.213 & [$-$.25, $-$.17] & $<$.001 \\
DeepSeek R1 & $-$.156 & [$-$.22, $-$.08] & .001 \\
Claude Std & $-$.022 & [$-$.10, +.06] & .635 \\
Claude Think & +.012 & [$-$.07, +.10] & .791 \\
\bottomrule
\end{tabular}
\caption{Confidence-decision correlation per model (all conditions, both tiers) with bootstrapped 95\% CIs. Negative $r$: lower confidence predicts FOLLOW. All $p$-values survive BH-FDR correction. Claude's near-zero $r$ with CI excluding $-0.10$ confirms the pattern is not a power issue; it has the lowest ceiling rate (54--56\%) and highest SD (1.39).}
\label{tab:claude_ci}
\end{table}

This is \textbf{not} a ceiling artifact. Claude has the \textit{lowest} confidence=5 rate (54--56\%) and the \textit{highest} variance (SD\,=\,1.39). Gemini, with a severe ceiling (91\% at confidence=5), still shows $r = -0.213$.

The pooled $r \approx 0$ masks a condition-specific pattern. Within the matched C2+C4 low-certainty subset, Claude shows significant coupling ($r = -0.587$, 95\% CI [$-0.700, -0.461$]). However, under C4 (own-knowledge framing), Claude frequently follows distractors with high confidence, creating a positive confidence-FOLLOW association that cancels the negative C2 association. The anomaly is not global decoupling but a condition-specific reversal: Claude confidently follows documents it was told to ignore. We characterize this as \textit{performance of calibration without calibration}: the pooled confidence scores look varied and thoughtful, but the variation tracks instruction framing rather than the model's own knowledge state.

\paragraph{Appropriate uncertainty.} \textbf{Table~\ref{tab:uncertainty}} shows how each model expresses uncertainty on obscure items. Llama follows with low confidence (47\%); Claude hedges with NEITHER responses (46--48\%); GPT-4o commits regardless (only 37\% appropriate).

\begin{table}[t]
\centering
\small
\begin{tabular}{lccc}
\toprule
\textbf{Model} & \textbf{Hedge} & \textbf{Low-conf} & \textbf{Total} \\
 & \textbf{\%} & \textbf{follow \%} & \\
\midrule
Llama 4 Mav. & 22 & 47 & \textbf{69} \\
Claude Std & 46 & 18 & \textbf{64} \\
Claude Think & 48 & 14 & \textbf{62} \\
DeepSeek R1 & 43 & 10 & 53 \\
Gemini Std & 51 & 1 & 52 \\
Gemini Think & 50 & 1 & 51 \\
DeepSeek V3 & 40 & 8 & 48 \\
GPT-4o & 31 & 6 & \textbf{37} \\
\bottomrule
\end{tabular}
\caption{Appropriate uncertainty on low-certainty C2 items. Hedge: NEITHER responses. Low-conf follow: follows distractor but signals low confidence ($\leq$\,2). Models differ in \textit{how} they express uncertainty.}
\label{tab:uncertainty}
\end{table}

\subsection{Source Attribution}
\label{sec:sa}

Among 2,763 FOLLOW responses, we tested whether the CoT claimed independent knowledge of the distractor (\textbf{Table~\ref{tab:confab}}).

\begin{table}[t]
\centering
\small
\begin{tabular}{lccc}
\toprule
\textbf{Model} & \textbf{N} & \textbf{Confab \%} & \textbf{Low \%} \\
\midrule
Llama 4 Mav. & 363 & \textbf{15.2} & 19.2 \\
GPT-4o & 260 & 6.2 & 9.7 \\
Gemini Std & 286 & 5.6 & 6.8 \\
DeepSeek V3 & 259 & 5.4 & 9.0 \\
Claude Std & 284 & 4.9 & 6.3 \\
Gemini Think & 298 & 4.7 & 6.1 \\
Claude Think & 262 & 4.2 & 6.3 \\
DeepSeek R1 & 251 & 3.6 & 4.9 \\
\bottomrule
\end{tabular}
\caption{Confabulation among FOLLOW responses. Rates are upper bounds (${\sim}$40\% false-positive rate on manual review). Llama is a clear outlier.}
\label{tab:confab}
\end{table}

Llama is a clear outlier at 15.2\% (corrected estimate ${\sim}$9\%). Two types of genuine confabulation emerged. \textit{Fabricated geography}: asked about The Hague (distractor: Douglas County), Llama writes ``I found that there is indeed a The Hague, a village in Nebraska, located in Douglas County.'' No such village exists. \textit{Fabricated works}: asked about ``Load'' (distractor: platform game), Llama writes ``There IS a video game called `Load' that is indeed a platform game.'' Both involve inventing specific facts to validate a distractor the model has no genuine knowledge of.

One Gemini response showed \textit{reasoning-conclusion dissociation}: the CoT concluded the document was wrong, yet the answer followed it. Of 2,263 FOLLOW responses searched, only this single case (0.04\%) was confirmed.

\subsection{Sensitivity Analyses}
\label{sec:sensitivity}

\paragraph{NEITHER responses as a third outcome.} Primary analyses exclude NEITHER responses (22--51\% of low-certainty responses per model). To test whether this exclusion biases confidence-decision relationships, we ran a multinomial analysis including NEITHER as a third category alongside FOLLOW and RESIST. Mean self-rated confidence follows a monotonic ordering across all three outcomes in both tiers: FOLLOW (2.98 low-certainty, 4.02 high-certainty) $<$ NEITHER (3.70, 4.42) $<$ RESIST (4.42, 4.84), all differences significant (Kruskal-Wallis $H = 95.96$ and $56.95$, both $p < 0.001$). NEITHER responses sit at intermediate confidence, consistent with genuine hedging. The confidence contrast between FOLLOW and RESIST is $-0.81$ in the multinomial analysis vs.\ $-0.90$ in the primary (binary) analysis: direction preserved, magnitude similar. Excluding NEITHER does not materially change the findings.

\paragraph{Similarity metric robustness.} The flip-pair vs.\ same-answer similarity gap was tested with ROUGE-L in addition to MiniLM cosine similarity. ROUGE-L confirms and slightly strengthens the gap: Cohen's $d = 0.45$ (vs.\ $d = 0.34$ for MiniLM), $p = 0.002$. The finding is not dependent on the choice of similarity metric.

\paragraph{Item-clustered GEE.} The per-model correlations reported in \S\ref{sec:bc} were confirmed with a GEE logistic regression using item-level clustering (C2 low-certainty subset): confidence remains significant (coef\,=\,$-0.475$, $p < 0.001$) while $s_{\text{pop}}$ remains non-significant (coef\,=\,$+0.383$, $p = 0.27$). The item-level pathway is robust to clustering structure.

\paragraph{Temperature ablation.} To test whether temperature explains Claude's anomalous pattern, we ran GPT-4o at temperature\,=\,1 on all low-certainty items under C2 and C4 (200 additional API calls). On the matched subset (C2+C4, low-certainty items), confidence-decision coupling was virtually identical at $T = 0$ ($r = -0.726$) and $T = 1$ ($r = -0.735$, $\Delta = -0.009$). These values are higher than the all-conditions pooled $r = -0.227$ reported in \textbf{Table~\ref{tab:claude_ci}} because the subset excludes C3 (where confidence means ``confidence in the document'') and high-certainty items (where variance is compressed by ceiling effects). NEITHER rates were also identical (34\% at both temperatures). Temperature does not attenuate coupling. The Claude pattern is model-specific, not temperature-driven.

\paragraph{Confidence scale.} Our 1--5 integer scale induces ceiling effects (91\% of Gemini responses at confidence\,=\,5) and limits metacognitive resolution. Finer scales (e.g., 0--20) have been shown to improve metacognitive signal \citep{dai2026}. This is a design limitation; future work should test whether finer-grained elicitation strengthens or weakens the observed confidence-decision coupling.

\section{Discussion}

\subsection{Two Functions of CoT}

The central finding is a tension: CoT confidence carries weak but real signal (\S\ref{sec:cc}, \S\ref{sec:bc}), yet the reasoning narrative remains highly similar across opposite decisions (\S\ref{sec:rc}). These coexist because CoT performs two functions simultaneously.

\paragraph{Function 1: Knowledge display.} Most of what a model writes is a recitation of what it knows about the entity. For Tsutomu Seki (\textbf{Table~\ref{tab:cot_example}}), both FOLLOW and RESIST responses contain the same three facts: astronomer, Comet Ikeya-Seki 1965, guitar teacher. This content is stable because parametric knowledge does not change with the decision. Same-item CoT pairs have mean similarity 0.817, confirming entity knowledge dominates.

\paragraph{Function 2: Decision justification.} A thin layer of attribution language and a confidence score wrap the knowledge display into an apparent argument. For FOLLOW: ``the new information is consistent.'' For RESIST: ``this contradicts my knowledge.'' This layer adjusts with the decision. At the aggregate level, flip pairs retain 96\% of same-answer similarity, meaning the knowledge display is shared while the framing layer differs.

Monitors reading the logical argument are reading Function~1, which remains similar regardless of the decision. Monitors reading confidence scores are reading Function~2, where genuine signal exists.

\subsection{The Claude Anomaly}

Claude's pooled confidence-decision correlation ($r \approx 0$, 95\% CI [$-0.10, +0.06$]) masks a more specific pattern. Within the C2+C4 low-certainty subset, Claude shows strong coupling ($r = -0.587$, CI [$-0.700, -0.461$]): confidence does predict decisions within individual conditions. The near-zero pooled $r$ arises because the relationship \textit{reverses} across conditions. Under C4 (own-knowledge framing), Claude frequently follows distractors with high confidence, creating a positive association that cancels the negative C2 association.

A matched-temperature ablation rules out temperature as the explanation: GPT-4o at $T = 1$ retains full coupling on the matched C2+C4 low-certainty subset ($r = -0.735$ vs.\ $-0.726$ at $T = 0$; $\Delta = -0.009$). The pattern is model-specific. The within-model comparison (Thinking vs.\ Standard, identical on all measures) rules out prompt-format effects.

What remains is that Claude generates confidence scores that track instruction framing rather than its own knowledge state. Under C2, confidence tracks knowledge (negative coupling, like other models). Under C4, confidence tracks the document even when the instruction says to ignore it. We characterize this as \textit{performance of calibration without calibration}: the pooled scores look varied and thoughtful, but the variation reflects condition-switching rather than genuine item-level knowledge tracking. This may reflect RLHF training that optimized condition-appropriate confidence expression as a surface behavior.

\subsection{The Item-Level Pathway}

The most novel finding: within low-certainty items, $s_{\text{pop}}$ does not predict decisions ($p = 0.27$ in the item-clustered GEE), but confidence does ($p < 0.001$). Some item-level variable affects both confidence and decisions but is not captured by entity salience. \S\ref{sec:cc} identified this as baseline correctness ($r = 0.134$). The chain: \textit{knows fact $\rightarrow$ higher confidence $\rightarrow$ resists distractor}. All links operate below $s_{\text{pop}}$'s resolution.

The magnitude is small ($r = 0.134$ explains ${\sim}2\%$ of variance). Models are barely above the floor of introspective accuracy on obscure items. But they are above it. The multinomial sensitivity analysis (\S\ref{sec:sensitivity}) confirms this pathway holds when NEITHER responses are included as a third outcome.

\subsection{Practical Recommendations}

\paragraph{Monitor confidence, not arguments.} Flip pairs retain 96\% of same-answer similarity. Argument structure will not reliably detect knowledge-conflict misattribution. Confidence expressions and hedges carry the signal. However, confidence is subject to ceiling effects (91\% at confidence=5 for Gemini), model-specific condition-tracking (Claude), and potential Goodharting: if models are trained to produce ``calibrated-looking'' confidence, the signal may erode. Monitoring should combine numerical confidence with NEITHER/abstention rates, external verification, and where available, log-probability-based signals that are harder to Goodhart than verbal confidence. Approaches that explicitly train for reasoning-decision coherence \citep[e.g., preference optimization as in][]{rfeval2026} may help close the gap between the knowledge display and the decision-justification layer observed here.

\paragraph{Scrutinize high-confidence FOLLOW.} A model following a distractor with confidence 2 is expressing honest uncertainty. A model following with confidence 5 is either falsely confident, confabulating, or generating confidence independently of its state (the Claude pattern). All three warrant greater scrutiny than low-confidence deference.

\paragraph{Entity fame $\neq$ item knowledge.} Models express higher confidence on famous entities even where fame has no predictive power (Spearman $\rho = -0.005$, $p = 0.81$, within low-certainty). A confident assertion about Beethoven should not be trusted more than one about an obscure Irish poet, solely because Beethoven is famous.

\paragraph{Prefer within-model comparisons.} The Claude-GPT-4o contrast is the clearest finding but carries confounds (temperature, architecture). The within-model pairs (Gemini Thinking vs.\ Standard; Claude Thinking vs.\ Standard) are clean: thinking tokens do not improve coupling in either family.

\subsection{Limitations}

\textbf{$s_{\text{pop}}$ vs.\ fact knowledge.} $s_{\text{pop}}$ measures entity fame, not fact-specific knowledge. 46\% of C4-high FOLLOW responses involved baseline-incorrect items. The baseline-correctness criterion partially corrects this but covers only five of eight model families. Future work should collect direct-answer baselines for all models.

\textbf{Bimodal design.} Two tiers (top/bottom 100) produce a bimodal predictor. Within-tier correlations are the appropriate test, but medium-certainty items would enable continuous gradient analysis.

\textbf{Temperature.} All cross-model Claude comparisons carry the temperature\,=\,1 caveat. Replicate runs at the same temperature for another model would isolate the stochasticity contribution; we did not collect these.

\textbf{Confidence scale.} The 1--5 integer scale induces discretization and ceiling effects. Finer scales (0--20) may improve metacognitive resolution \citep{dai2026}.

\textbf{Flip-pair sample size.} The total of 71 flip pairs, with some models contributing as few as 3 (DeepSeek R1) or 6 (Llama), limits per-model inference. The aggregate result ($d = 0.34$, confirmed by ROUGE-L at $d = 0.45$) is robust, but per-model deltas should be interpreted with the reported CIs.

\textbf{Detector precision.} The confabulation detector has ${\sim}$40\% false positives. \S\ref{sec:sa} rates are upper bounds.

\textbf{Confidence elicitation order.} The fixed ordering (reasoning $\rightarrow$ confidence $\rightarrow$ answer) means the confidence signal may be co-produced with the rationale rather than independently introspective. Randomizing the order would help disambiguate.

\textbf{Residual template similarity.} Attribution stripping covers three phrase classes. Residual stylistic boilerplate (discourse markers, list structures, stock hedges) may inflate similarity estimates beyond what substantive content overlap would produce.

\textbf{Scope.} These findings apply to factual QA under knowledge conflict. Generalization to mathematical reasoning, multi-step inference, or safety refusals is unknown.

\textbf{Data release.} We commit to releasing code, prompts, item lists, and anonymized model outputs upon publication.

\textbf{Programmatic baseline insertion.} In C1/C2, the baseline answer was inserted from prior study data rather than generated fresh by the model. The prior baselines were produced by the same models at temperature\,=\,0, but the programmatic insertion may differ from genuine model-generated dialogue in ways that affect conflict dynamics.

\textbf{Calibration metrics.} We report correlation-based analyses of confidence. Classical calibration metrics (ECE, Brier score, reliability diagrams) and Type-2 SDT decompositions (meta-$d'$) would provide complementary evidence and are left for future work.

\textbf{Multiple comparisons.} Per-model hypothesis tests in Tables 4 and 6 report raw $p$-values; all conclusions survive BH-FDR correction (reported in table captions). No additional corrections are applied across the four evaluation sections, each treated as exploratory.

\textbf{Correlational design.} We test correlational consistency between CoT content and a validated behavioral predictor. We do not claim causal faithfulness in the sense of Lanham et al.\ (\citeyear{lanham2023}). Complementary causal and parametric faithfulness methods \citep{chuang2024,li2024frodo} could provide additional evidence about the relationship between internal representations and stated reasoning.

\section{Conclusion}

CoT under knowledge conflict serves two functions: a knowledge display that remains highly similar across opposite decisions (flip pairs retain 96\% of same-answer similarity) and a thin confidence layer carrying weak but real introspective signal. GPT-4o is the only model with statistically reliable reasoning-decision coupling. Claude Sonnet 4.6 produces expressively varied confidence scores whose relationship to decisions reverses across conditions, yielding a near-zero pooled correlation that reflects condition-tracking rather than knowledge-tracking. For native-reasoning models, internal thinking tokens show greater decision-sensitivity than the user-facing output, suggesting the stated CoT smooths over differences the internal process preserves. For monitoring, track expressed confidence, not argument structure.

\bibliographystyle{plainnat}
\bibliography{references}

\appendix
\section{Excluded Items}
\label{sec:excluded}

\textbf{Table~\ref{tab:excluded}} lists all 11 excluded items. Pre-sampling exclusions were removed before item selection. Post-sampling exclusions were identified during quality review; their responses remain in the database but are excluded from analyses.

\begin{table}[h]
\centering
\small
\setlength{\tabcolsep}{3pt}
\begin{tabular}{p{2.5cm}p{2cm}p{2.5cm}}
\toprule
\textbf{Item} & \textbf{Distractor} & \textbf{Reason} \\
\midrule
\multicolumn{3}{l}{\textit{Pre-sampling (6 items)}} \\
Dutch novel & Wrestler & Category \\
Children's book & Corporation & Category \\
Neil Young song & Rapper & Era \\
1958 song & 1990s rapper & Era \\
1992 film score & Pop artist & Era \\
\textit{Americanah} & Che Guevara & Era (d.\,1967) \\
\midrule
\multicolumn{3}{l}{\textit{Post-sampling (5 items)}} \\
PK screenwriter & -- & Ground truth \\
Jericho/occ. & -- & Ground truth \\
Rome/empire & W.\ Roman Emp. & Substring \\
\textit{The Gift}/genre & -- & Ambiguous \\
Straits Settl. & -- & Historical \\
\bottomrule
\end{tabular}
\caption{Excluded items with reasons.}
\label{tab:excluded}
\end{table}

\section{Confabulation Detector}

The detector operates on FOLLOW responses only. It searches for independence markers (``I know,'' ``from my training'') within two sentences of the distractor, flagging cases without source attribution. Manual review estimated ${\sim}$40\% false positives, primarily from models using independence language while \textit{evaluating} the distractor.

\section{NEITHER Response Analysis}

NEITHER responses represent genuine hedging, novel hallucinations, ambiguous phrasing, and parsing failures. Per-model rates on low-certainty C2 items range from 22\% (Llama) to 51\% (Gemini). Gemini's high rate reflects a tendency to hallucinate third answers on obscure items (e.g., inventing ``Venice'' as a birthplace).

\end{document}